\documentclass[nohyperref]{article}

\usepackage{microtype}
\usepackage{graphicx}
\usepackage{subcaption} % לפאנלים עם כיתובים (a), (b)
\usepackage{booktabs,siunitx} % for professional tables
\usepackage{hyperref}

\usepackage[accepted]{icml2022}

\usepackage{amsmath}
\usepackage{amssymb}
\usepackage{mathtools}
\usepackage{amsthm}
\DeclareMathOperator{\CAA}{CAA}
\newcommand{\CAAmax}{\CAA_{\max}}
\DeclareMathOperator{\Var}{Var}

\DeclareMathOperator{\lcm}{lcm}
\usepackage[capitalize,noabbrev]{cleveref}

\theoremstyle{plain}
\newtheorem{theorem}{Theorem}[section]
\newtheorem{proposition}[theorem]{Proposition}
\newtheorem{lemma}[theorem]{Lemma}

\theoremstyle{definition}
\newtheorem{definition}[theorem]{Definition}

\theoremstyle{remark}
\newtheorem{remark}[theorem]{Remark}

\usepackage[textsize=tiny]{todonotes}

\icmltitlerunning{Submission and Formatting Instructions for ICML 2022}
\begin{document}
\twocolumn[
\icmltitle{Complexity as advantage}
\icmlsetsymbol{equal}{*}

\begin{icmlauthorlist}
\icmlauthor{Oshri Naparstek}{yyy}

\end{icmlauthorlist}

\icmlaffiliation{yyy}{IBM Research}

\icmlkeywords{Complexity, Advantage}

\vskip 0.3in
]
%\begin{abstract}
%What makes data complex in a way that matters for learning? Classical notions---entropy, Kolmogorov complexity, logical depth---capture facets of structure, but either blur noise and rich regularities or remain uncomputable in practice. We introduce \textbf{Complexity-as-Advantage (CAA)}: a source appears complex when some observers consistently outperform others on it. Formally, CAA is defined as the dispersion of predictive regrets relative to the best observer in a reference family. This observer-relative lens can be connected to classical quantities---under log-loss, CAA gaps correspond to conditional mutual information atoms that sum to excess entropy; along a compute budget ladder, CAA profiles resemble logical depth; and when observers are compressors, CAA relates to Kolmogorov and MDL complexity through excess description length. 

%Empirically, CAA exhibits the expected U-shape on tunable sources, highlights hidden seasonality in time series, distinguishes structured ciphertext from random noise, and surfaces long-range dependencies in natural language. These preliminary results suggest that CAA may serve as a practical framework for probing usable structure, offering a complementary tool for analyzing datasets and models in modern machine learning.
%\end{abstract}

\begin{abstract}
What makes complexity \emph{useful}? 
We introduce \textbf{Complexity-as-Advantage (CAA)}, an operational framework that defines complexity as the dispersion of regret across a family of resource-bounded observers. 
Under log-loss and Markov ladders, adjacent CAA gaps coincide with conditional mutual-information atoms, and their sum recovers \emph{excess entropy}. 
This link grounds CAA in statistical decision theory and information theory, while a complementary coding view interprets it as the variance of excess description length under MDL. 
Empirically, CAA profiles yield scalar indicators of \emph{logical depth}---such as the fraction of tail advantage or budget thresholds---that cleanly separate shallow, chaotic, and deep processes in cellular automata and cryptographic tasks. 
Together, these results position CAA as a unifying, computable lens on data complexity: it explains when additional computational investment yields genuine predictive advantage, and why structure invisible to single observers becomes measurable across a hierarchy of capabilities.
\end{abstract}

\section{Introduction}

Why do large language models easily capture the rhythms of Shakespeare yet fail on random noise, even though both defeat a standard compressor such as \texttt{gzip}? The difference is not raw entropy but \emph{usable structure}: Shakespeare contains patterns that stronger observers can exploit, while noise does not. Yet classical complexity measures—entropy, Kolmogorov complexity, logical depth—either conflate these cases or remain uncomputable for resource-bounded observers.

We address this gap by introducing \textbf{Complexity-as-Advantage (CAA)}, an operational framework that defines complexity through the \emph{performance gaps} it induces across a family of observers. Complexity is high when better observers consistently outperform weaker ones. This reframes complexity from an abstract property of a source into a practical diagnostic: it asks when and where additional computational investment yields genuine predictive advantage.

CAA connects several classical notions under one roof. Under log-loss and Markov ladders, CAA gaps coincide with conditional mutual-information atoms and their sum recovers excess entropy. Along compute–budget ladders, CAA produces \emph{advantage profiles} that provide scalar indicators of logical depth, separating shallow, chaotic, and deep processes. When observers are compressors, CAA aligns with the MDL principle by interpreting complexity as the variance of excess description length. These links anchor CAA simultaneously in decision theory, information theory, and coding.

Empirically, we estimate CAA from regret curves and show that it uncovers distinctions invisible to entropy or compression alone. On synthetic sources, CAA differentiates periodic-but-noisy signals from pure noise. In cryptographic ladders, it exposes concentrated structure missed by single-observer metrics. In cellular automata, it yields depth indicators that classify rules 90, 30, and 110 into shallow, chaotic, and deep regimes.

\paragraph{Contributions.}
This paper makes four main contributions:
\begin{itemize}
    \item We propose \textbf{CAA}, a general framework that defines complexity as the dispersion of regret across resource-bounded observers.
    \item We show that CAA provides an \emph{operational view of logical depth}: advantage profiles along a compute-budget ladder yield scalar depth indicators (e.g., tail fraction, budget thresholds) that empirically distinguish shallow, chaotic, and deep processes.
    \item We establish theoretical links: under log-loss and Markov ladders, CAA gaps coincide with conditional mutual-information atoms and their sum recovers excess entropy; a complementary coding view interprets CAA as the variance of excess description length (MDL).
    \item We validate CAA through controlled experiments, including cellular automata and cryptographic tasks, showing that it exposes structured complexity invisible to single-observer measures.
\end{itemize}

Taken together, these results position CAA as a unifying and computable lens on data complexity: a framework that connects decision theory, information theory, and machine learning, while offering practical diagnostics for when structure is exploitable.

\section{Related Work}

The definition of complexity has been pursued from multiple angles, spanning decision theory, information theory, and learning theory. Our contribution is to unify these perspectives under a single operational lens: \emph{Complexity-as-Advantage (CAA)}, which reframes complexity as the dispersion of regret across resource-bounded observers.

\paragraph{Decision Theory and Regret.}
Classical statistical decision theory, following Wald and Savage, employs regret as a prescriptive criterion for optimality \cite{Savage1954}. Recent work in online learning studies variance-aware regret bounds and model selection for a \emph{single} learner \cite{Mukherjee2018, Dann2024, Jia2024}. CAA departs from both: it treats regret not as a tool for policy selection, but as a \emph{descriptive measure of environmental structure}, elevating regret dispersion itself as the object of study.

\paragraph{Algorithmic and Statistical Information Theory.}
Kolmogorov and Chaitin defined complexity via minimal description length \cite{Kolmogorov1965, Chaitin1969}, while MDL refines this into a coding principle for model selection \cite{Grunwald2007}. Bennett’s logical depth \cite{Bennett1988} and Gell-Mann’s effective complexity \cite{GellMann1996} aim to capture “meaningful” structure beyond randomness. These notions are powerful yet either uncomputable or agnostic to resources. CAA operationalizes them: it provides a relativistic, budget-dependent criterion that yields computable depth indicators distinguishing shallow, chaotic, and deep processes.

\paragraph{Predictive Information and Excess Entropy.}
Computational mechanics formalizes statistical complexity and excess entropy as measures of total predictable information \cite{Crutchfield1989, Shalizi2001, Bialek2001}. Related work connects predictive information to causal structure via rate–distortion style objectives \cite{StillCrutchfield2010}. CAA refines this viewpoint by decomposing excess entropy into \emph{observer-dependent advantage gaps}, revealing structural heterogeneity invisible to a single scalar entropy.

\paragraph{Complexity in Machine Learning.}
Modern ML offers parallel lenses on complexity. Dataset difficulty and scaling laws quantify performance growth with data and compute \cite{Kaplan2020, Hernandez2020, Sorscher2023}. Intrinsic-motivation RL uses curiosity bonuses as proxies for learnability \cite{Pathak2017, burda2019}. Representation-focused views relate generalization to compression and information in the weights \cite{Arora2018, Achille2019}. CAA complements these by offering an observer-agnostic criterion for \emph{where} gains are possible: environments that yield systematic regret dispersion across computational budgets.

\paragraph{Summary.}
Prior work highlights different slices of the complexity puzzle: regret minimization, description length, predictive information, and empirical scaling. \textbf{CAA unifies these threads into a single operational framework, defining complexity as measurable advantage dispersion and grounding abstract notions of depth in concrete, computable indicators.}

\section{Complexity-as-Advantage}

\subsection{General Definition}

\begin{definition}[Asymptotic Average Loss]
Let $X=(X_u)_{u\in\mathcal{I}}$ be a process indexed by a countable set 
$\mathcal{I}$ (e.g.\ time steps, spatial locations, or nodes in a graph).  
Given a predictor $A$ producing predictions $\hat{y}^A_u$ and a loss function $\ell$, define
\begin{equation}
\label{eq:def-L}
\begin{split}
L(A;X)\;\triangleq\;
\limsup_{|\Lambda|\to\infty}\;\frac{1}{|\Lambda|}\;
\sum_{u\in\Lambda} \ell\!\big(\hat{y}^A_u, X_u\big),
\end{split}
\end{equation}
where $\Lambda\subset\mathcal{I}$ ranges over an increasing sequence of finite regions.  
The minimal achievable loss and the regret are
\begin{align}
L^\ast(X) &\triangleq \inf_{A\in\mathcal{A}} L(A;X),
\label{eq:def-Lstar}\\
R(A;X)    &\triangleq L(A;X)-L^\ast(X).
\label{eq:def-R}
\end{align}
\end{definition}

\begin{definition}[Complexity-as-Advantage (CAA)]
Given a reference distribution $\pi$ on $\mathcal{A}$,
\begin{align}
\CAA(X;\mathcal{A},\pi)
&\triangleq \Var_{A\sim\pi}\!\big[R(A;X)\big],
\label{eq:def-CAA}
\end{align}
with a \emph{gap} variant
\begin{align}
\CAA_{\max}(X)
&\triangleq \sup_{A,B\in\mathcal{A}} \big|R(A;X)-R(B;X)\big|.
\label{eq:def-CAAmax}
\end{align}
\end{definition}

\begin{lemma}[Two-Algorithm Closed Form]
\label{lem:two-alg}
Let $\mathcal{A}=\{A_{\text{naive}},A_{\text{soph}}\}$ with
$L(A_{\text{soph}};X)\le L(A_{\text{naive}};X)$. 
Set $\Delta L \triangleq L(A_{\text{naive}};X)-L(A_{\text{soph}};X)\ge 0$
and let $\pi(A_{\text{naive}})=p$. Then
\begin{align}
\CAA(X;\mathcal{A},\pi) &= p(1-p)\,(\Delta L)^2,
\label{eq:twoalg-var}\\
\CAA_{\max}(X) &= \Delta L.
\label{eq:twoalg-gap}
\end{align}
In particular, for the uniform prior $p=\tfrac12$, 
\begin{equation}
\CAA(X)=\tfrac14\,\big(\CAA_{\max}(X)\big)^2.
\label{eq:twoalg-uniform}
\end{equation}
\end{lemma}

\begin{proof}
Here $R(A_{\text{soph}};X)=0$ and $R(A_{\text{naive}};X)=\Delta L$,
so $R$ is a two-point random variable with support $\{0,\Delta L\}$ and probabilities $\{1-p,p\}$.  
Its variance is $p(1-p)(\Delta L)^2$, giving \eqref{eq:twoalg-var}.  
The gap is $\sup_{A,B}|R(A;X)-R(B;X)|=\Delta L$, yielding \eqref{eq:twoalg-gap}.
\end{proof}

\begin{remark}
This formulation does not assume $X$ is stochastic or time-indexed.  
It applies equally to deterministic or stochastic processes, and to different
domains such as time series, spatial images, or general index sets.  
\end{remark}

\subsection{Specialization: Log-Loss and Markov Predictors}

We now consider the important special case of log-loss,
\begin{equation}
\ell(x,\hat P)=-\log_2 \hat P(x),
\label{eq:def-logloss}
\end{equation}
which ties prediction performance directly to information-theoretic quantities.

\begin{definition}[Markov predictors under log-loss]
For a stationary process $X=(X_t)$, the order-$m$ Markov predictor $A^{(m)}$
achieves
\begin{equation}
\label{eq:logloss-entropy}
L\big(A^{(m)};X\big)= H\!\left(X_t \,\middle|\, X_{t-1},\dots,X_{t-m}\right),
\end{equation}
with the convention $L(A^{(0)};X)=H(X_t)$.
\end{definition}

\begin{proposition}[Adjacent Markov orders]
\label{prop:adjacent}
For $m\ge 1$,
\begin{equation}
\label{eq:adjacent-gap}
\Delta L_m \;\triangleq\; L\big(A^{(m-1)};X\big)-L\big(A^{(m)};X\big)
= I\!\left(X_t;X_{t-m}\,\middle|\, X_{t-1}^{\,t-m+1}\right),
\end{equation}
so the gap $\CAA_{\max}$ between order-$m$ and order-$(m{-}1)$
equals this conditional mutual information.
\end{proposition}
\begin{theorem}[CAA gaps and Excess Entropy]
\label{thm:caa-excess}
Let $E$ denote the excess entropy of a stationary process,
\begin{equation}
E\;\triangleq\;I\!\big(X_{-\infty}^{t-1};X_t\big).
\label{eq:def-excess}
\end{equation}
Then the cumulative CAA gaps telescope into the predictive information:
\begin{align}
\sum_{m=1}^{M} \Delta L_m
&= H(X_t) - H\!\left(X_t \,\middle|\, X_{t-1}^{\,t-M}\right).
\label{eq:telescoping}
\end{align}
Taking $M\to\infty$ yields
\begin{equation}
\lim_{M\to\infty} \sum_{m=1}^{M} \Delta L_m \;=\; 
H(X_t)-H\!\left(X_t\,\middle|\,X_{-\infty}^{t-1}\right) \;=\; E.
\label{eq:sum-to-E}
\end{equation}
In particular, if $X$ is a finite-order $K$ Markov process, the sum truncates exactly at $m=K$:
\begin{equation}
E \;=\; H(X_t)-H\!\left(X_t\,\middle|\,X_{t-1}^{\,t-K}\right).
\label{eq:finite-K}
\end{equation}
\end{theorem}

\begin{proof}[Proof sketch]
\textit{Step 1} (log-loss $\Rightarrow$ conditional entropy):  
From \eqref{eq:logloss-entropy}, the $m$-step observer has loss  
$L\big(A^{(m)};X\big)=H\!\left(X_t\middle|X_{t-1}^{\,t-m}\right)$.  

\noindent
\textit{Step 2} (telescoping):  
Subtracting consecutive losses gives
\begin{align*}
\Delta L_m &= H\!\left(X_t\middle|X_{t-1}^{\,t-m+1}\right)
             - H\!\left(X_t\middle|X_{t-1}^{\,t-m}\right) \\
           &= I\!\left(X_t; X_{t-m}\,\middle|\,X_{t-1}^{\,t-m+1}\right).
\end{align*}
Summing up to horizon $M$ yields \eqref{eq:telescoping}.  

\noindent
\textit{Step 3} ($M\to\infty$):  
In the limit, the conditioning expands to the entire past, giving \eqref{eq:sum-to-E}.  
For a $K$-order Markov process, conditioning stabilizes at $M=K$, giving \eqref{eq:finite-K}.
\end{proof}

\begin{remark}
CAA gaps decompose the predictive information into conditional MI atoms, one for each new rung of context.  
Thus CAA offers an \emph{operational} view of excess entropy: each $\Delta L_m$ is the realized advantage from extending context by one step, and the total advantage budget coincides with $E$.  
For finite-order Markov sources this identity is exact and finite; for general processes it holds as a convergent series.  
The same reasoning extends to spatial or graph-indexed processes by replacing temporal lags with expanding neighborhoods.
\end{remark}
\begin{remark}[Generalization and practical observers]
The derivation above uses the one-step definition 
$E = I(X_{-\infty}^{t-1};X_t)$. 
More generally, excess entropy is defined as
\[
E = I\!\big(X_{-\infty}^{t-1}; X_{t:\infty}\big),
\]
the mutual information between the entire past and entire future. 
CAA admits a corresponding two-dimensional decomposition:
each $\Delta L_{m,k}$ quantifies the predictive gain at horizon $k$
from extending context by $m$ steps, and the double sum
recovers $E$. 

Moreover, the equalities here assume an idealized family of observers
that can achieve the entropy bounds. For any restricted observer class
(e.g.\ bounded-memory predictors or finite neural models), the cumulative
CAA advantage is always \emph{upper-bounded} by $E$, 
with equality only for the omniscient observer.
This makes CAA both a decomposition of predictive information in theory
and a practical lower estimate of excess entropy in applied settings.
\end{remark}

\section{Empirical demonstrations}

We present two complementary experiments. The first uses a tunable source to
exhibit the characteristic U-shape of our complexity–as–advantage ($\CAA$)
measure as structure varies from pure noise to perfect order. The second
shows \emph{relativistic} complexity: the advantage depends on the observer
family, with a strong separation on a cryptographic source.

\subsection{Experiment I: A tunable source and the U-curve}

\paragraph{Source.}
For $p\in[0,1]$, define a binary process $X^{(p)}=(X_t)$ as a Bernoulli
mixture between a deterministic periodic base and white noise. At time $t$,
with probability $p$ emit the next symbol from a fixed periodic template;
with probability $1-p$ emit a fair coin. We use two templates:
(i) period-$2$ pattern $[0,1]$; (ii) period-$6$ pattern $[0,0,0,1,1,1]$.

\paragraph{Observers and loss.}
Under log-loss $\ell(x,\hat P)=-\log_2\hat P(x)$ we evaluate two online
Markov predictors: a naive order-$k_{\mathrm{n}}$ and a sophisticated
order-$k_{\mathrm{s}}$ (Laplace smoothing). The asymptotic average loss
$L(A;X)$ is estimated by the average online log-loss over length $N$.
For two observers the performance gap is
$\Delta L \triangleq L(A_{\mathrm{naive}};X)-L(A_{\mathrm{soph}};X)$,
and with a uniform prior on $\{A_{\mathrm{naive}},A_{\mathrm{soph}}\}$ we have
\begin{equation}
\CAA(X)\;=\;\tfrac{1}{4}\,(\Delta L)^2.
\label{eq:caa-twoalg}
\end{equation}

\paragraph{Protocol.}
For each $p$ we generate $B$ i.i.d.\ sequences (random template phase per
sequence), compute $L$ for each observer, and report the mean and standard
deviation of $\Delta L$ and $\CAA$ across the $B$ runs. We use
$(N,B,\alpha)=(6\cdot 10^4,\,16,\,1.0)$ and two pairs:
\noindent\textbf{Pair A:} $(k_{\mathrm{n}},k_{\mathrm{s}})=(1,3)$ on period-2.\\
\textbf{Pair B:} $(k_{\mathrm{n}},k_{\mathrm{s}})=(3,5)$ on period-6.

\paragraph{Findings.}
Pair~A is almost monotone in $p$: for $p=1$ order-$1$ cannot lock the
template phase, so $L(A^{(1)};X)$ remains high while $L(A^{(3)};X)$ drops,
yielding a large gap. Pair~B shows the predicted U-curve: the gap (and thus
$\CAA$ via Eq.~\eqref{eq:caa-twoalg}) is small at $p\approx 0$ (white noise)
and $p\approx 1$ (both orders suffice for the clean period-$6$ signal), and
maximizes at intermediate $p$.

\begin{figure}[t]
  \centering
  \includegraphics[width=.48\textwidth]{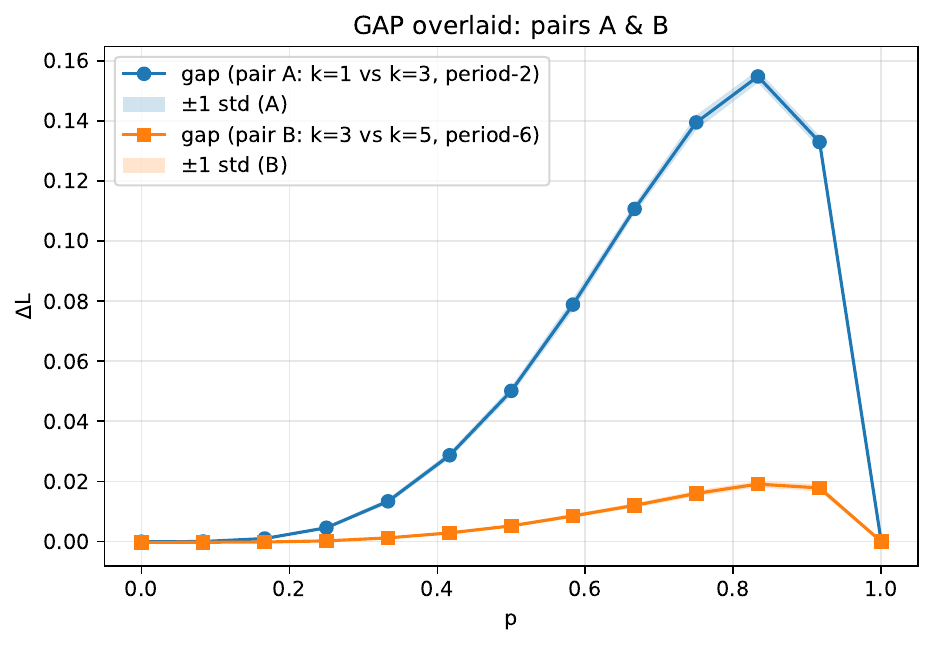}
  \caption{Gap $\Delta L$ versus $p$ for both observer pairs. Pair A
  (period-$2$, orders $1$ vs.\ $3$) is mostly monotone; Pair B (period-$6$,
  orders $3$ vs.\ $5$) shows a clear U-shape. Shaded bands: mean$\pm$std over $B$ sequences.}
  \label{fig:u-gap}
\end{figure}

\begin{figure}[t]
  \centering
  \includegraphics[width=.48\textwidth]{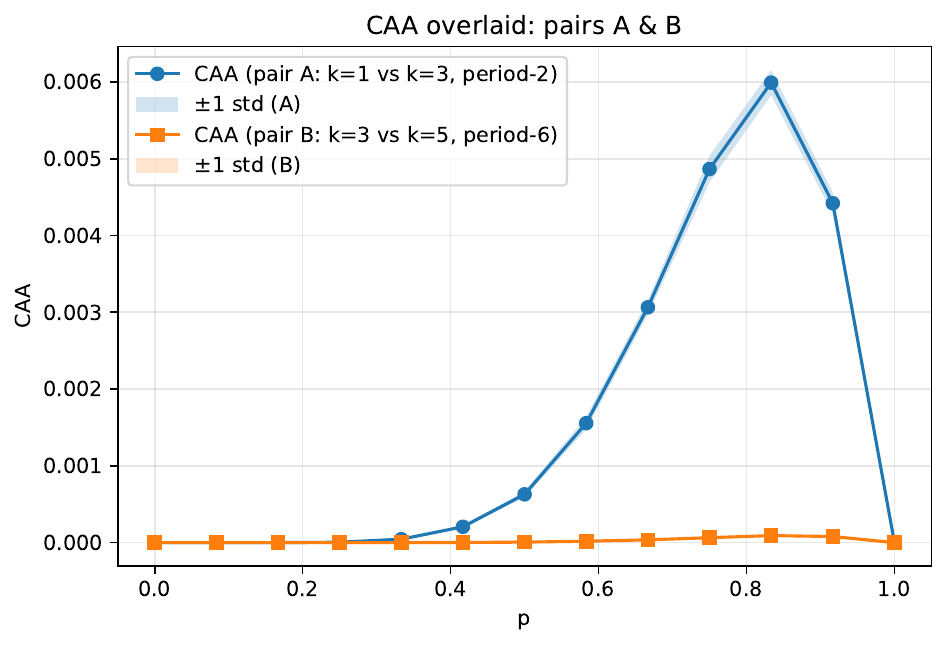}
  \caption{$\CAA$ versus $p$ (uniform prior); corresponds to
  Fig.~\ref{fig:u-gap} via Eq.~\eqref{eq:caa-twoalg}.}
  \label{fig:u-caa}
\end{figure}

\subsection{Experiment II: Relativistic complexity on statistical vs.\ cryptographic sources}

\paragraph{Sources.}
We contrast two stationary binary processes: (i) a two-state HMM with sticky
transitions and biased emissions, and (ii) a ``crypto'' source obtained by
XOR-ing the alternating plaintext $(0,1,0,1,\ldots)$ with a repeating key of
length $m$ (after a random prefix and a plaintext key reveal for alignment).
For the HMM we use
\begin{subequations}\label{eq:hmm-params}
\begin{align}
T&=\begin{bmatrix}0.98&0.02\\ 0.02&0.98\end{bmatrix}, \label{eq:hmm-T}\\
E&=\begin{bmatrix}0.85&0.15\\ 0.15&0.85\end{bmatrix}. \label{eq:hmm-E}
\end{align}
\end{subequations}
The ciphertext of the crypto source is periodic with period $\lcm(2,m)$; for
large $m$ (or a pseudo-random keystream) it appears nearly i.i.d.\ to
low-order statistics.

\paragraph{Observer families.}
\emph{Statistical} observers are order-$k$ Markov models. \emph{Search}
observers are XOR-seekers that, given the key, decrypt, predict a
deterministic alternating plaintext bit, and re-encrypt (without a key they
fall back to a Markov model). We evaluate four cases:
\[
\text{Stat/Stat},\quad \text{Stat/Search},\quad
\text{Crypto/Stat},\quad \text{Crypto/Search}.
\]

\paragraph{Protocol and metrics.}
We measure online average log-loss after a burn-in. For crypto, the burn-in
aligns with the start of encryption (prefix $+$ key). For Search, we reset
the internal key phase so the first scored prediction is in the correctly
decrypted coordinate system. We report $\CAAmax=\Delta L$ (gap) per case.

\paragraph{Findings.}
\emph{Stat/Stat} and \emph{Stat/Search} are essentially identical on the HMM,
as Search without a key reduces to a Markov model. On crypto we see a strong
separation: \emph{Crypto/Search} attains a near-maximal gap (decryption
collapses uncertainty, so $L_{\text{soph}}\approx 0$), whereas
\emph{Crypto/Stat} is small with long or pseudo-random keys (ciphertext
near i.i.d.). With a short periodic key, \emph{Crypto/Stat} can be nonzero
due to the $\lcm(2,m)$ aliasing: higher-order Markov models partially lock
onto the induced periodicity while low-order models do not.

\begin{figure}[t]
  \centering
  \includegraphics[width=.4\textwidth]{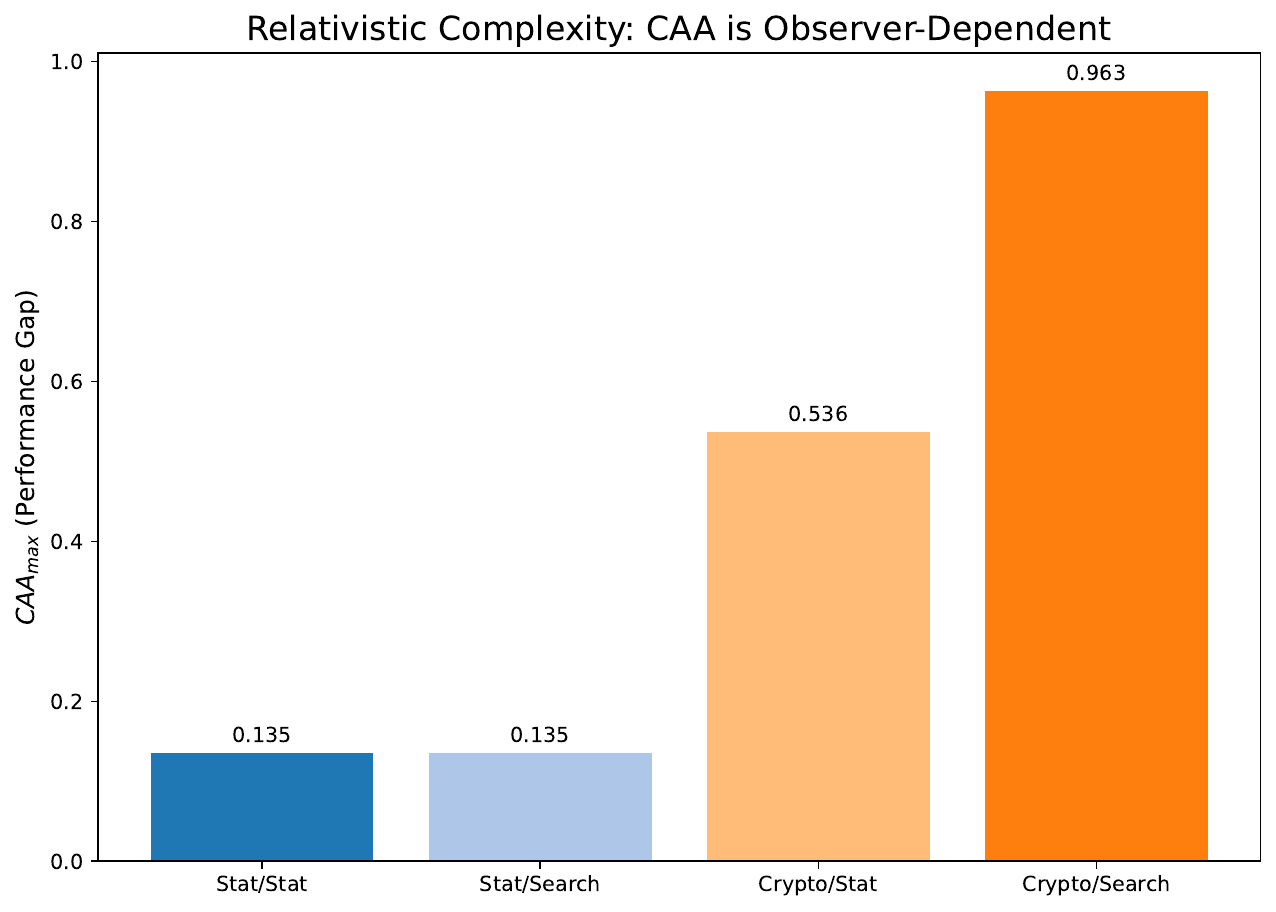}
  \caption{Relativistic complexity: $\CAAmax$ (gap) for four source$\times$observer
  combinations. Crypto/Search is high because decryption collapses uncertainty;
  Crypto/Stat is small with long or pseudo-random keys.}
  \label{fig:relativistic}
\end{figure}

\paragraph{Reproducibility.}
Figures are vector PDFs. Defaults: sequence length $N=5\!\times\!10^4$;
U-curve ensemble size $B=16$; Laplace smoothing $\alpha=1$. Changing $m$
explains the \emph{Crypto/Stat} bar: short periodic keys induce learnable
periodicity; long or pseudo-random keys suppress it.

\section{Computational Budgets and Logical Depth (CAA view)}

Classical notions of \emph{logical depth} (Bennett, 1988) describe how much
computation is required to ``unfold'' the structure of a source.  
Within the CAA framework we can make this idea \emph{operational}: 
logical depth is simply the \emph{budget-indexed advantage profile}
of a fixed observer family $\{A^{(b)}\}$ as the compute budget $b$ grows. 
In other words, depth is what remains hidden until observers invest 
substantial computational resources.

\subsection{Definition: ladders of observers}

Let $\{A^{(b)} : b=0,1,2,\dots\}$ be an observer family at increasing budgets
(e.g.\ search depth, rollout length, or observation radius).
Define the incremental improvement
\begin{align}
\Delta L_b \;\triangleq\; L\!\big(A^{(b-1)};X\big) \;-\; L\!\big(A^{(b)};X\big) \;\ge 0.
\end{align}
Each $\Delta L_b$ is a two-rung CAA gap.
Shallow processes concentrate their gains early (large $\Delta L_b$ at small $b$),
while deep processes defer gains to large $b$.
Thus the profile $\{\Delta L_b\}$ is the CAA view of logical depth.

\subsection{Case study 1: a crypto ladder with a decisive spike}

\textbf{Setup.}  
We construct a simple cryptographic source: a repeating key XOR applied
to a $0101\dots$ plaintext. Observer $A^{(b)}$ is a key-searcher restricted
to key lengths $\ell \leq b$.

\begin{figure}[t]
  \centering
  \begin{subfigure}{.49\linewidth}
    \includegraphics[width=\linewidth]{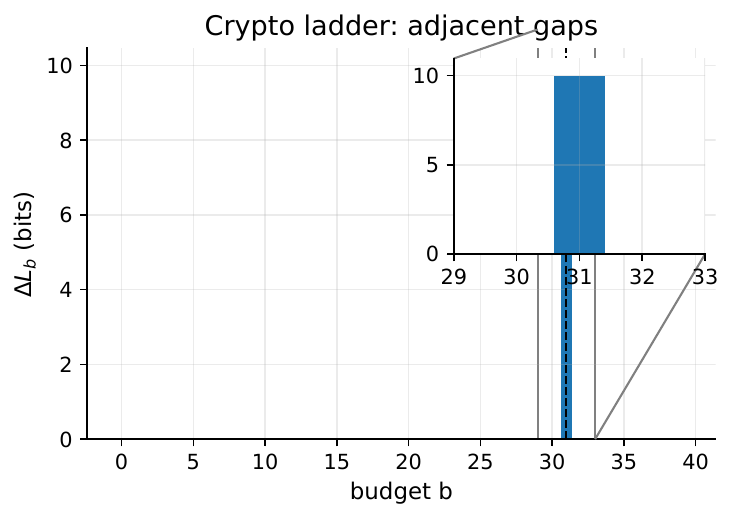}
    \caption{Incremental gains $\Delta L_b$ vs.\ budget $b$. A single spike at $b^\star$.}
  \end{subfigure}\hfill
  \begin{subfigure}{.49\linewidth}
    \includegraphics[width=\linewidth]{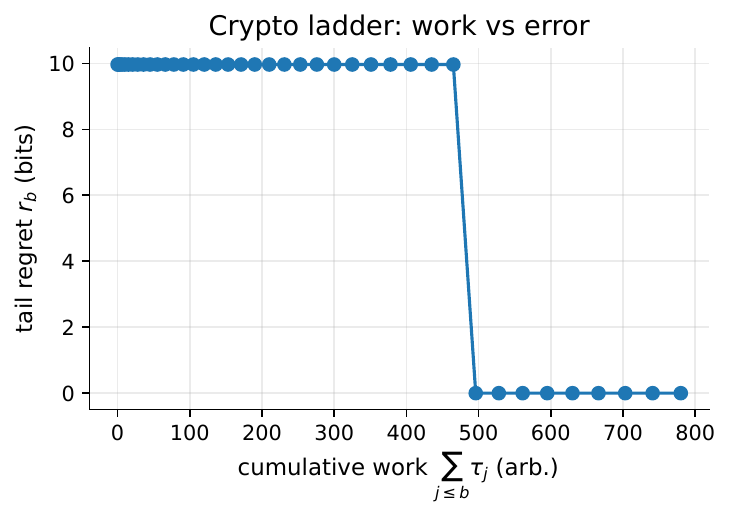}
    \caption{Tail regret $r_b$ vs.\ cumulative work. The collapse at $b^\star$ highlights a depth barrier.}
  \end{subfigure}
  \caption{\textbf{Crypto ladder.} CAA captures a textbook depth barrier: all advantage is locked until one decisive budget threshold.}
  \label{fig:crypto-depth-clean}
\end{figure}

\textbf{Finding.}  
Figure~\ref{fig:crypto-depth-clean} shows a \emph{sharp spike} in $\Delta L_b$
precisely at the true key length $b^\star$.
All usable advantage concentrates at one budget threshold,
and the tail regret $r_b$ collapses there.
This is a canonical depth signature: no incremental improvements,
but a decisive gain once the critical budget is crossed.

\subsection{Case study 2: cellular automata ladders}

\textbf{Setup.}  
We compare three canonical one-dimensional automata:
Rule~90 (additive, shallow), Rule~30 (chaotic), and Rule~110 (complex).
Observer $A^{(r)}$ is a local simulator with radius $r$ (budget).

\begin{figure}[t]
  \centering
  \includegraphics[width=\linewidth]{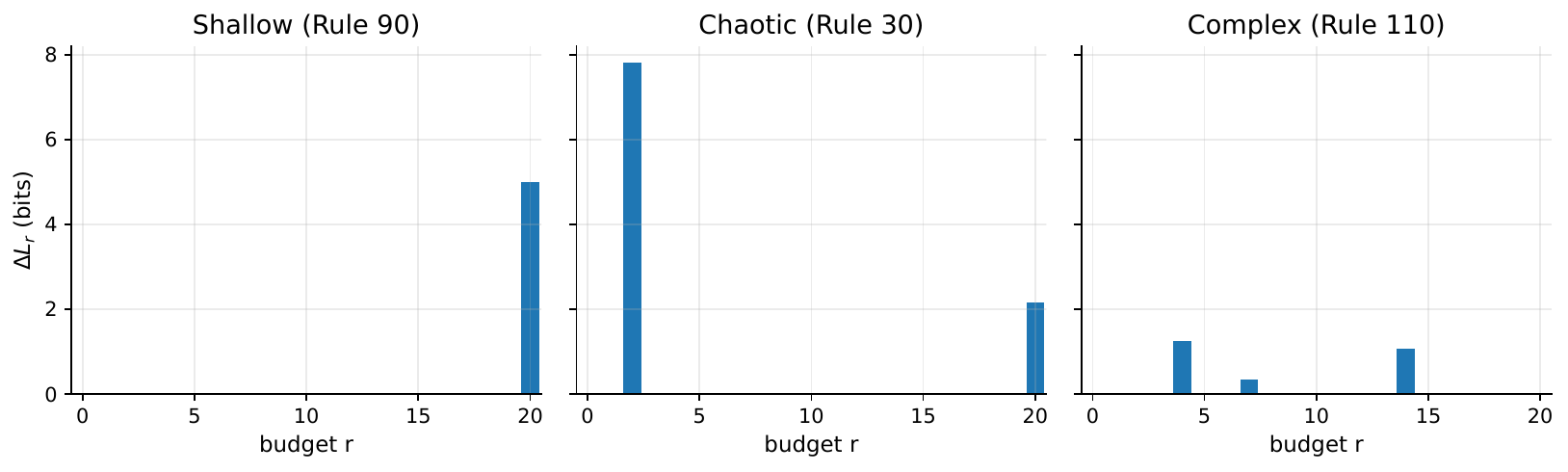}
  \caption{\textbf{CA ladders ($k=20$).} Rule~90 (shallow) gains early, Rule~30 (chaotic) gains diffusely, Rule~110 (deep) defers improvements to large budgets.}
  \label{fig:ca-triptych-clean}
\end{figure}

\textbf{Finding.}  
Figure~\ref{fig:ca-triptych-clean} shows the $\Delta L_r$ profiles ($k=20$).
- Rule~90 is front-loaded: gains appear at small $r$ and then vanish.  
- Rule~30 yields only weak, diffuse improvements: no budget helps much.  
- Rule~110 is tail-heavy: significant gains emerge only at large $r$.  

\begin{figure}[t]
  \centering
  \includegraphics[width=.85\linewidth]{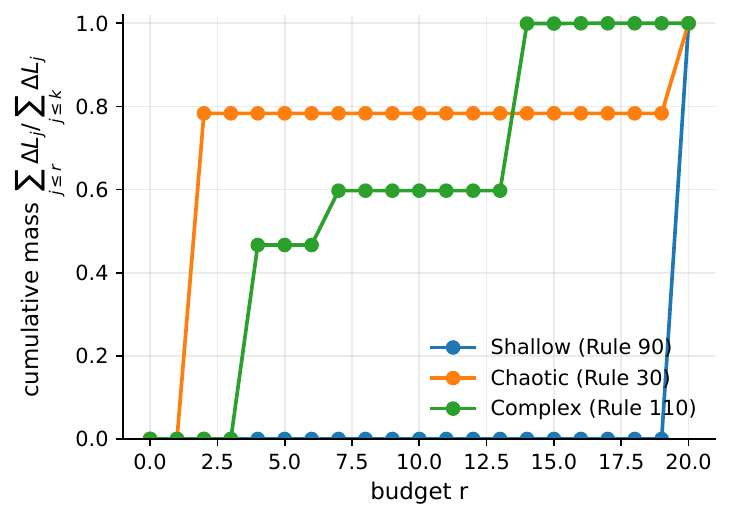}
  \caption{\textbf{Cumulative mass curves.} Shallow processes rise early; deep processes defer gains, producing late-rise profiles.}
  \label{fig:ca-cdf-clean}
\end{figure}

Figure~\ref{fig:ca-cdf-clean} condenses this: shallow processes rise early,
while deep ones defer mass to late budgets.

\subsection{Scalar indicators of depth}

The full $\{\Delta L_b\}$ profile can be summarized by simple scalars:

\paragraph{Tail fraction.}  
\begin{align}
\mathrm{TailFrac}_\alpha \;=\; \frac{\sum_{j>\lfloor \alpha B\rfloor}\Delta L_j}{\sum_{j=1}^{B}\Delta L_j}.
\end{align}
Large values indicate tail-heavy (deep) structure.

\paragraph{Half-mass budget.}  
\begin{align}
b_{50} \;=\; \min\{\, b:\; \sum_{j\leq b}\Delta L_j \geq M/2 \,\}.
\end{align}
Deep processes exhibit larger $b_{50}$.

\paragraph{Normalized depth score.}  
\begin{align}
D \;=\; \frac{1}{B}\cdot \frac{\sum_{b=1}^{B} b\,\Delta L_b}{\sum_{b=1}^{B}\Delta L_b}.
\end{align}
This is the average relative budget at which gains occur.

\paragraph{Empirical summary.}  
Table~\ref{tab:depth-metrics} reports these indicators.
Rule~90 is shallow (all gains upfront), Rule~30 is chaotic (little gain anywhere),
and Rule~110 is deep (meaningful gains only at large budgets).

\begin{table}[t]
  \centering
  \caption{\textbf{Scalar depth indicators (CA, $k=20$).} CAA converts depth into measurable scalars. Rule~90: trivial and shallow. Rule~30: chaotic, negligible gains. Rule~110: deep, with deferred structure.}
  \label{tab:depth-metrics}
  \begin{tabular}{lccc}
\toprule
Process & TailFrac$_{2/3}$ & $b_{50}$ & $D$ \\
\midrule
Shallow (Rule 90) & 1.00 & 20 & 1.00 \\
Chaotic (Rule 30) & 0.22 & 2 & 0.29 \\
Complex (Rule 110) & 0.40 & 7 & 0.42 \\
\bottomrule
\end{tabular}
 
\end{table}

\subsection{Takeaway}

Under CAA, logical depth becomes measurable. 
Shallow processes expose their structure immediately,
chaotic ones yield no exploitable advantage,
and deep ones defer gains to high-budget observers.  
This distinction holds across natural budget ladders 
(search depth, rollout length, observation radius),
positioning CAA as a \emph{practical diagnostic for depth}.

\section{Kolmogorov-Style Complexity as \texorpdfstring{$\mathrm{CAA}$}{CAA}}
\label{sec:kolmogorov-caa}

\subsection{Setup: description length as loss}
Let $x^{n}=(x_{1},\dots,x_{n})$ be a finite sample from a source $X$.
For a lossless coder $A$, let $L_{n}(A;x^{n})$ be its codelength on $x^{n}$,
with per-symbol average
\begin{align}
  \label{eq:avg-code-length}
  \bar L_{n}(A;x^{n}) \;\triangleq\; \tfrac{1}{n}\,L_{n}(A;x^{n}).
\end{align}
For probabilistic coders this equals the empirical log-loss,
so \emph{description length and predictive loss coincide}.
The asymptotic expected loss is
\begin{align}
  L(A;X) \;\triangleq\; \limsup_{n\to\infty}\, \mathbb{E}[\bar L_{n}(A;X^{n})].
\end{align}

Given a class $\mathcal A$ of coders,
\begin{align}
  L^{\ast}(X) &\;\triangleq\; \inf_{A\in\mathcal A} L(A;X), &
  R(A;X) &\;\triangleq\; L(A;X)-L^{\ast}(X).
\end{align}
CAA is then defined as the \emph{dispersion of regret}:
\begin{align}
  \mathrm{CAA}(X;\mathcal A,\pi)
  \;\triangleq\; \mathrm{Var}_{A\sim\pi}[R(A;X)].
\end{align}
Large CAA means some coders incur much more excess length than others—clear
evidence of exploitable structure.

\subsection{Link to Kolmogorov and MDL}
Kolmogorov complexity $K(x^{n})$ is the length of the shortest program for $x^{n}$,
with practical coders $A$ giving upper bounds
$K(x^{n}) \le L_{n}(A;x^{n})+O(1)$.  
The gap
\begin{align}
  R_{n}(A;x^{n})
  \;=\; \bar L_{n}(A;x^{n}) - \min_{B\in\mathcal A}\bar L_{n}(B;x^{n})
\end{align}
is the \emph{excess description length}.  
CAA therefore measures how spread out these excess lengths are across coders:
a source has high CAA if \emph{different coding strategies succeed very differently}.  
This reframes KC/MDL from “absolute complexity” to “advantage potential.”

\paragraph{Extremes.}
- If all coders are asymptotically optimal for the same class, regrets coincide
and $\mathrm{CAA}=0$.  
- For i.i.d.\ noise, all coders converge to the entropy rate—again $\mathrm{CAA}\approx 0$.  
- Nonzero CAA arises only when \emph{some coders exploit structure that others cannot.}

\subsection{Practical estimation}
Empirically we estimate per-symbol lengths, subtract the best performer,
and compute either variance or max-gap. Averaging across many sequences yields
a stable CAA estimate.  
This makes CAA a \emph{directly measurable quantity}, not just a theoretical construct.

\subsection{Observer dependence: a simple experiment}
\begin{figure}[t]
  \centering
  \includegraphics[width=\linewidth]{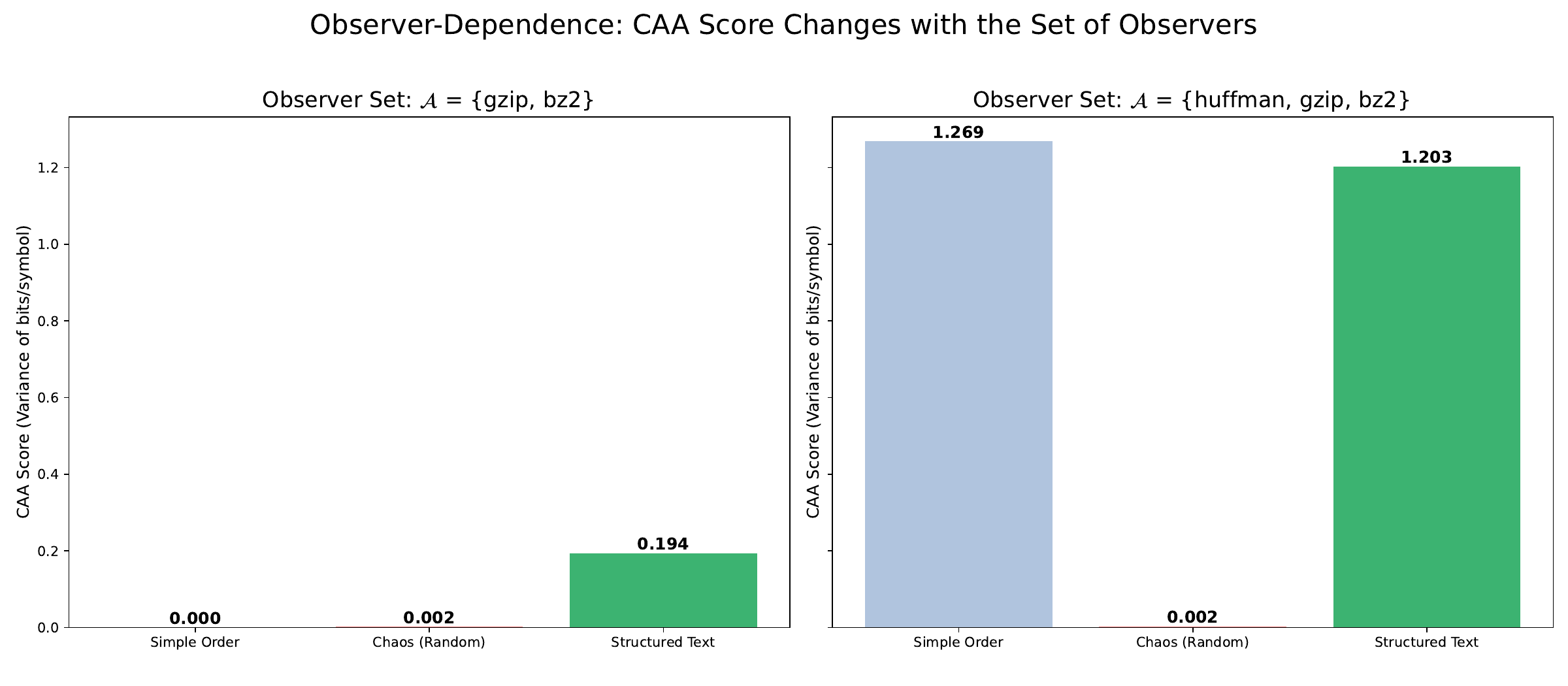}
  \caption{\textbf{Observer dependence.} Adding Huffman to the observer set
  surfaces advantage gaps: CAA increases for sources with dependencies beyond $H_{0}$
  (periodicity, text), but remains near zero for pure noise.}
  \label{fig:observer-dependence}
\end{figure}

We compared three sources—periodic strings, i.i.d.\ noise, and English text—
under two observer sets:
\[
\mathcal A_{1}=\{\texttt{gzip},\texttt{bz2}\}, \quad
\mathcal A_{2}=\{\texttt{huffman},\texttt{gzip},\texttt{bz2}\}.
\]
Results (Fig.~\ref{fig:observer-dependence}, Table~\ref{tab:caa-observers}):
- With $\mathcal A_{1}$, CAA $\approx 0$ for pure order and pure noise,
modest for text.  
- Adding Huffman (\(\mathcal A_{2}\)) makes CAA \emph{jump sharply}
for periodic and text, but not noise.  

\emph{Why?} Huffman captures only zeroth-order frequencies,
while LZ coders exploit longer dependencies.  
Thus periodicity and text create a clear advantage gap between observers,
and CAA detects it immediately.

\begin{table}[t]
  \centering
  \caption{\textbf{CAA as variance of excess codelength (bits/symbol).} 
  Adding Huffman reveals hidden structure: CAA increases for order and text, 
  but not for noise.}
  \label{tab:caa-observers}
  \begin{tabular}{l
                  S[table-format=1.3]
                  S[table-format=1.3]
                  S[table-format=+1.3]}
    \toprule
    \textbf{Source} &
    {$\mathcal{A}_1$ }&
    {$\mathcal{A}_2$} &
    {$\Delta$ (A2--A1)} \\
    \midrule
    Simple order      & 0.000 & 1.269 & +1.269 \\
    Chaos (i.i.d.)    & 0.002 & 0.002 &  0.000 \\
    Structured text   & 0.194 & 1.203 & +1.009 \\
    \bottomrule
  \end{tabular}
\end{table}

\subsection{Discussion and caveats}
\paragraph{What CAA is (and is not).}
CAA does not assert an absolute “true complexity.”  
It is observer-relative: it measures the spread in excess description length
\emph{for the chosen observer set}. This is its strength—it diagnoses where
advantage gaps exist.

\paragraph{Choice of prior and robustness.}
The prior $\pi$ controls which gaps matter more.  
Practical stabilizers: (i) priors aligned with compute budgets,
(ii) trimmed variance (drop outliers) to check robustness.

\paragraph{Controls.}
- Block-shuffling that destroys long-range dependencies collapses Huffman–LZ gaps,
reducing CAA as expected.  
- Adding a run-length encoder closes the gap on periodic strings, again lowering CAA.  
These controls confirm that CAA is sensitive to \emph{which capabilities are included.}

\paragraph{Finite-sample issues.}
Short sequences add header/warmup overheads; we mitigate with long samples,
overhead correction, and averaging across draws.

\paragraph{Takeaway.}
In the Kolmogorov/MDL setting, CAA is simply the variance (or gap) of
\emph{excess codelengths} across coders.  
It is low when observers are equally powerless or equally strong,
and high exactly when structure exists that only some observers can exploit.
\section{Discussion and Broader Impact}

CAA provides a unifying lens on complexity that is directly relevant to
machine learning. While our experiments used synthetic sources and
classical coders, the principle extends naturally:

- \textbf{Dataset difficulty.} Scaling laws in deep learning
\cite{Kaplan2020,Hernandez2020} quantify performance as a function of data
and model size, but lack a structural criterion. CAA explains why
performance gaps arise: some datasets contain exploitable patterns that
weaker models miss, producing high advantage dispersion.

- \textbf{Inductive bias.} Different architectures embody different
observer families. CAA formalizes when an inductive bias matters: a bias
is useful exactly when it yields a lower regret than alternatives,
increasing the spread.

- \textbf{Intrinsic motivation.} Curiosity-driven RL heuristics reward
agents for surprise \cite{Pathak2017}. CAA grounds this intuition in
information theory: states with high advantage potential are precisely
those where stronger observers outperform weaker ones.

These connections suggest that CAA is not only a theoretical construct but
a practical diagnostic: a tool for identifying when and where learning
capacity pays off. Future work should test this on modern neural predictors
and large-scale datasets, but the framework is in place.
\section{Conclusion}

We introduced \emph{Complexity-as-Advantage (CAA)}, a framework that
recasts complexity as the dispersion of predictive regret across a family
of observers. CAA makes classical notions operational: it connects to
Kolmogorov and MDL via excess description length, to Bennett’s logical
depth via budget-indexed advantage profiles, and to excess entropy via
theoretical identities. 

Empirically, CAA distinguishes shallow, chaotic, and deep processes:
periodic order is trivial, chaos offers no usable advantage, and complex
sources like Rule~110 or natural text defer their gains to higher-budget
observers. 

By framing complexity as advantage, CAA bridges decision theory,
algorithmic information, and learning theory. We believe it can serve as a
diagnostic tool for machine learning, clarifying which datasets contain
exploitable structure and which models can capitalize on it.

\bibliography{bibiliography}
\bibliographystyle{icml2022}

\end{document}